\documentclass{SCIS2026}

\usepackage{amsmath,amsfonts,bm}

\def\eqref#1{equation~\ref{#1}}
\def\1{\bm{1}}

\DeclareMathAlphabet{\mathsfit}{\encodingdefault}{\sfdefault}{m}{sl}
\SetMathAlphabet{\mathsfit}{bold}{\encodingdefault}{\sfdefault}{bx}{n}

\usepackage{algorithm}
\usepackage{float}
\usepackage{algpseudocode}
\usepackage{placeins}
\definecolor{promptheader}{RGB}{221,232,244}
\definecolor{promptshared}{RGB}{236,244,251}
\definecolor{promptnumerical}{RGB}{244,245,247}
\definecolor{promptcontext}{RGB}{252,246,231}

\begin{document}
\raggedbottom

% Journal metadata. Complete the empty fields after acceptance.
\ArticleType{RESEARCH PAPER}
\Year{2026}
\Month{}
\Vol{}
\No{}
\DOI{}
\ArtNo{}
\ReceiveDate{}
\ReviseDate{}
\AcceptDate{}
\OnlineDate{}
\AuthorMark{Liu Z D}
\AuthorCitation{Liu Z D, Yang J Q, Cheng M Y, Li Z, Chen E H}

\title{World2TS: Real-World Event Context for Time-Series Forecasting}{World2TS: Real-World Event Context for Time-Series Forecasting}

% SCIS convention: \dag marks equal contribution and * marks correspondence.
% The three corresponding authors share the contact email below.
\author[1\dag]{Zhiding Liu}{}
\author[2\dag]{Jiqian Yang}{}
\author[1*]{Mingyue Cheng}{}
\author[2]{Zhi Li}{}
\author[1*]{Enhong Chen}{{cheneh@ustc.edu.cn}}
\contributions{These authors contributed equally to this work.}
\address[1]{State Key Laboratory of Cognitive Intelligence, University of Science and Technology of China, China}
\address[2]{Shenzhen International Graduate School, Tsinghua University, China}

\abstract{Time-series forecasting plays a vital role in a wide range of real-world applications. Recent approaches have achieved strong performance by modeling temporal correlations and cross-channel dependencies, yet most forecasting benchmarks still represent time series primarily as numerical histories. In practice, time series are observations of complex real-world processes, and their trajectories may reflect events, interventions, disruptions, and changes in measurement or reporting that cannot always be inferred from past values alone. A central barrier to studying context-aware forecasting is therefore the lack of large-scale datasets that align real-world events with the individual series they affect. We present \textbf{World2TS}, a scalable construction framework and large-scale dataset for time-series forecasting with real-world event context. We first develop \texttt{world2ts-core}, a reusable tool that combines agent-led semantic reasoning and Web search, command-line execution and validation, and human review within a feedback-driven discovery workflow. Given heterogeneous time series with semantic descriptions, the workflow discovers candidate events, consolidates retrieved evidence into verified events, and links them to relevant series under explicit temporal and semantic scopes. Using this framework, we construct \texttt{world2ts-data}, which contains 31 datasets across six domains, comprising 4,179 series, 97.1 million observations, 1,159 events, and 56,846 event–series links. We further conduct extensive data analyses and controlled comparative experiments to characterize the dataset and demonstrate the utility of the collected context for improving forecasting accuracy. World2TS provides a reusable data foundation for developing and evaluating context-aware forecasting models and agents. The project homepage is available at \url{https://world2ts.github.io/}.}


\maketitle

\section{Introduction}
\label{sec:introduction}
Time-series forecasting is a fundamental task in time-series analysis and underpins decision-making in applications such as traffic management, financial forecasting, healthcare analytics, and weather prediction~\cite{jin2023spatio,li2024master,kaushik2020ai,bi2023pangu}.
It has consequently attracted sustained research attention~\cite{qiu2024tfb,cheng2025comprehensive}.
Deep learning has substantially advanced the field through architectures that capture temporal dependencies~\cite{zhou2021informer,zhou2022fedformer} and cross-channel dependencies~\cite{zhang2023crossformer,liu2024itransformer,liu2026disents}.
Further progress has come from inductive biases and training strategies designed for non-stationary data~\cite{kim2022revin,liu2023san}, as well as foundation models pretrained across large and diverse time-series corpora~\cite{das2024timesfm,liu2024timer,woo2024moirai}.
Despite this architectural progress, most methods still formulate forecasting primarily as a numerical pattern-learning problem: predictions are derived from historical values and, when available, numerical covariates, while the real-world circumstances that generate those values are rarely represented explicitly.

However, \textbf{time series are observations of complex real-world processes rather than isolated numerical sequences}.
Their trajectories can reflect external events, interventions, disruptions, and changes in how a process is measured or reported, whose implications may be difficult to infer from historical patterns alone.
Consider a public-health surveillance series after its reporting agency changes from daily updates to a weekly schedule: delayed counts and periodic spikes may resemble a change in disease dynamics even when they primarily reflect a new observation process.
The numerical history alone cannot reliably distinguish these explanations, whereas the reporting announcement makes the change interpretable and can alter the resulting forecast.
This limitation has motivated pioneer efforts toward context-aware forecasting~\cite{jia2024gpt4mts,wang2024newsforecast}.
Time-MMD aligns numerical series with textual information collected from real-world sources across multiple domains~\cite{liu2024timemmd}, while CiK constructs forecasting tasks in which carefully designed textual context is essential for accurate prediction~\cite{williams2025context}.
Together, these studies demonstrate the promise of supplementing numerical observations with language-accessible context.

Nevertheless, existing resources are built around fixed combinations of numerical and textual sources or task-specific authored contexts, limiting their scalability and ability to fully reflect the correlation between time series data and the real world. Therefore, it remains an open problem to construct a general framework that starts from heterogeneous time series, discovers relevant real-world events through adaptive web search, verifies the supporting evidence, and determines the particular series to which each event applies.
Scalability in this setting is fundamentally a semantic challenge.
Different datasets observe different physical, social, or economic processes, while channels within the same dataset may represent distinct variables, entities, and locations. Consequently, a scalable construction framework must flexibly interpret heterogeneous series, formulate domain- and time-specific searches, reconcile dispersed evidence, and ground each event in the appropriate temporal and semantic scope.

To address this problem, we introduce \textbf{World2TS}, a flexible and scalable construction framework and large-scale dataset for time-series forecasting with real-world event context.
Its reusable tool, \texttt{world2ts-core}, combines agent-led semantic reasoning and web search, command-line execution and validation, and human review within a feedback-driven discovery workflow.
Given heterogeneous time series with semantic descriptions, the workflow discovers candidate events, consolidates retrieved evidence into verified events, and links them to relevant series under explicit temporal and semantic scopes.
Using this framework, we construct \texttt{world2ts-data}, which contains 31 datasets across energy, transport, weather, health, finance, and cryptocurrency, comprising 4,179 series, 97.1 million observations, 1,159 verified events, and 56,846 event--series links.
We conduct extensive data analyses and controlled comparative experiments to characterize the resulting resource, assess the reliability of its event layer, and demonstrate the utility of the collected context for improving forecasting accuracy.

Our contributions are threefold:
\begin{itemize}
    \item \textbf{A scalable construction framework.}
    We develop \texttt{world2ts-core}, a reusable Agent--CLI--Human cooperative workflow that combines feedback-driven event discovery, strict execution and validation, and explicit series-association rules to flexibly augment heterogeneous time-series collections.
    \item \textbf{A large-scale real-world event dataset.}
    We construct \texttt{world2ts-data}, a cross-domain resource containing 31 datasets, 4,179 series, 97.1 million observations, 1,159 verified events, and 56,846 fine-grained event--series links.
    \item \textbf{A systematic validation of the resulting context.}
    Through extensive dataset analyses and controlled forecasting comparisons, we assess the characteristics, reliability, and predictive utility of the constructed event layer, establishing World2TS as a reusable data foundation for developing and evaluating context-aware forecasting models and agents.
\end{itemize}

\section{Related Work}
\label{sec:related_work}
\paragraph{Numerical time-series forecasting.}
Time-series forecasting has developed over several decades, with early approaches drawing on statistical theory to formulate predictive models. ARIMA combines differencing to address non-stationarity with a linear model of past observations and past forecast errors~\cite{hyndman2021forecasting}. Although effective for capturing linear temporal structure, this formulation has limited ability to represent nonlinear dynamics~\cite{zhang2003hybridarima}.
With the continued development of deep learning, neural models have become widely used for time-series forecasting and have given rise to diverse architectural paradigms for modeling temporal dependencies. Convolution-based models, such as ConvTimeNet and TimesNet, respectively use hierarchical convolution and two-dimensional convolutional blocks to capture local patterns and multi-period temporal variation~\cite{cheng2025convtimenet,wu2023timesnet}. MLP-based models, such as DLinear, decompose series into trend and seasonal components and learn separate linear mappings for each~\cite{zeng2023transformers}. Attention-based models capture long-range temporal dependencies and cross-variable relationships~\cite{zhou2021informer,liu2024itransformer}. Moreover, recent work has explored unified forecasting with long-context Transformers. Timer-XL generalizes next-token prediction to multivariate time series and models varying-length contexts, covering univariate, multivariate, and covariate-informed forecasting~\cite{liu2025timerxl}. These deep learning models have substantially advanced time-series forecasting by enabling more flexible modeling of complex temporal dynamics.

Despite their increased modeling capacity, most deep forecasters are trained on a single dataset, limiting the diversity of patterns observed during training and their generalizability across domains. Time-series foundation models instead leverage large-scale pretraining on heterogeneous collections to enable zero-shot forecasting. Early efforts primarily adopt two formulations. Continuous-valued sequence models learn transferable representations directly from numerical series~\cite{das2024timesfm,woo2024moirai,rasul2023lag,taga2025timepfn}. Discrete language models quantize time-series values into tokens for probabilistic forecasting~\cite{ansari2024chronos}.

\paragraph{Context-aware forecasting.}
In addition to modeling the target history, several studies incorporate information about the surrounding environment. In pioneering works, this information is represented as numerical covariates, aligned with the target series, and explicitly modeled to predict future values~\cite{das2023long,olivares2023neural}. TimeXer uses cross-attention to integrate exogenous variables with the target series~\cite{wang2024timexer}. Such approaches can account for external factors that are observed as structured measurements, but they require the relevant information to be available as numerical series. Many real-world drivers of temporal change, including policy decisions, disruptions, and market events, are instead documented in unstructured text. However, conventional time series models inherently struggle to ingest such non-numerical signals. To bridge this gap, the emergence of large language models (LLMs) has expanded context modeling from numerical covariates to textual evidence. For instance, GPT4MTS collects related textual information and integrates it with numerical histories through prompting~\cite{jia2024gpt4mts}.
Once text is introduced as context, the challenge extends beyond multimodal fusion to determining which evidence is relevant and how it should affect the forecast. News-to-Forecast addresses this issue through iterative news selection and reflection on potential event impacts~\cite{wang2024newsforecast}.Recent work further casts forecasting as an agentic workflow rather than a single prediction step~\cite{garza2025timecopilot,zhao2025timeseriesscientist}. CastFSR further combines forecaster selection with contextual retrieval and reflective refinement~\cite{tao2026castfsr}.

\paragraph{Time-series datasets and benchmarks.}
Time-series datasets and benchmarks have evolved in response to the changing data requirements of forecasting models. Earlier methods were commonly trained and evaluated on individual datasets, making it difficult to determine whether their performance transferred across application domains. As forecasting models began to learn from collections of related series, the Monash Time Series Forecasting Archive consolidated numerical datasets with diverse domains, sampling frequencies, series lengths, and missing-data characteristics into a shared resource for standardized evaluation~\cite{godahewa2021monash}. The emergence of time-series foundation models further shifted the focus from within-dataset fitting to large-scale pretraining and zero-shot generalization, creating a need for both broader training corpora and more comprehensive evaluation. GIFT-Eval addresses this need by providing a non-leaking pretraining collection of approximately 230 billion observations together with a zero-shot evaluation suite comprising 23 datasets, more than 144,000 series, and 177 million observations across seven domains and ten sampling frequencies~\cite{aksu2024gifteval}. This progression expands benchmark from comparing forecasting accuracy on shared numerical datasets to assessing whether pretraining produces cross-domain generalization.

Beyond numerical series, the growing use of multimodal and context-aware forecasting models has created a need for resources that connect temporal dynamics with real-world evidence, such as reports, news, and event descriptions. To bridge this modality gap, Time-MMD establishes multimodal alignments across nine domains, offering a comprehensive real-world testbed for text-attributed forecasting~\cite{liu2024timemmd}. However, text--series alignment alone does not reveal whether the supplied context is necessary for prediction. CiK therefore constructs controlled forecasting tasks in which synthetic contextual information is essential to making the forecast~\cite{williams2025context}. More recently, TimesX has moved toward richer and refreshable contextual resources by covering 190 variables from 19 domains and automatically collecting and verifying metadata, calendar information, related variables, and events~\cite{liu2026timesx}. This evolution, from text-aligned collections to context-essential tasks and automatically maintained real-world resources, expands benchmark design toward evaluating both the use and the availability of external evidence.

Despite these advances, existing benchmarks provide context for predefined collections. World2TS instead addresses the construction of reusable event context for a new, heterogeneous time-series collection. Its Agent--CLI--Human workflow discovers and verifies real-world events within dataset-specific temporal and semantic scopes, then grounds them to relevant series, producing verified events detail text and reusable event--series links.

% Organize as three compact paragraphs:
% (1) Time-series datasets and benchmarks.
% (2) Context-aware and multimodal forecasting: Time-MMD, Time-IMM, CiK, and TimesX.
% (3) Agent-assisted data curation, including TimeSeriesExamAgent.
% Table: context, series-level alignment, construction extensibility, human review,
% and evaluation coverage, with consistent definitions across resources.
% Add verified BibTeX entries when drafting the corresponding discussion.

\section{World2TS-Core: Collaborative Dataset Construction}
\label{sec:construction}

\begin{figure}[t]
    \centering
    % Replace the placeholder below with the workflow figure from the project page.
    % Original asset: https://world2ts.github.io/world2ts-assets/framework.pdf?v=20260905
    % After uploading the PDF, replace the \fbox block with:
    % \includegraphics[width=\linewidth]{figures/framework.pdf}
   \includegraphics[width=\linewidth]{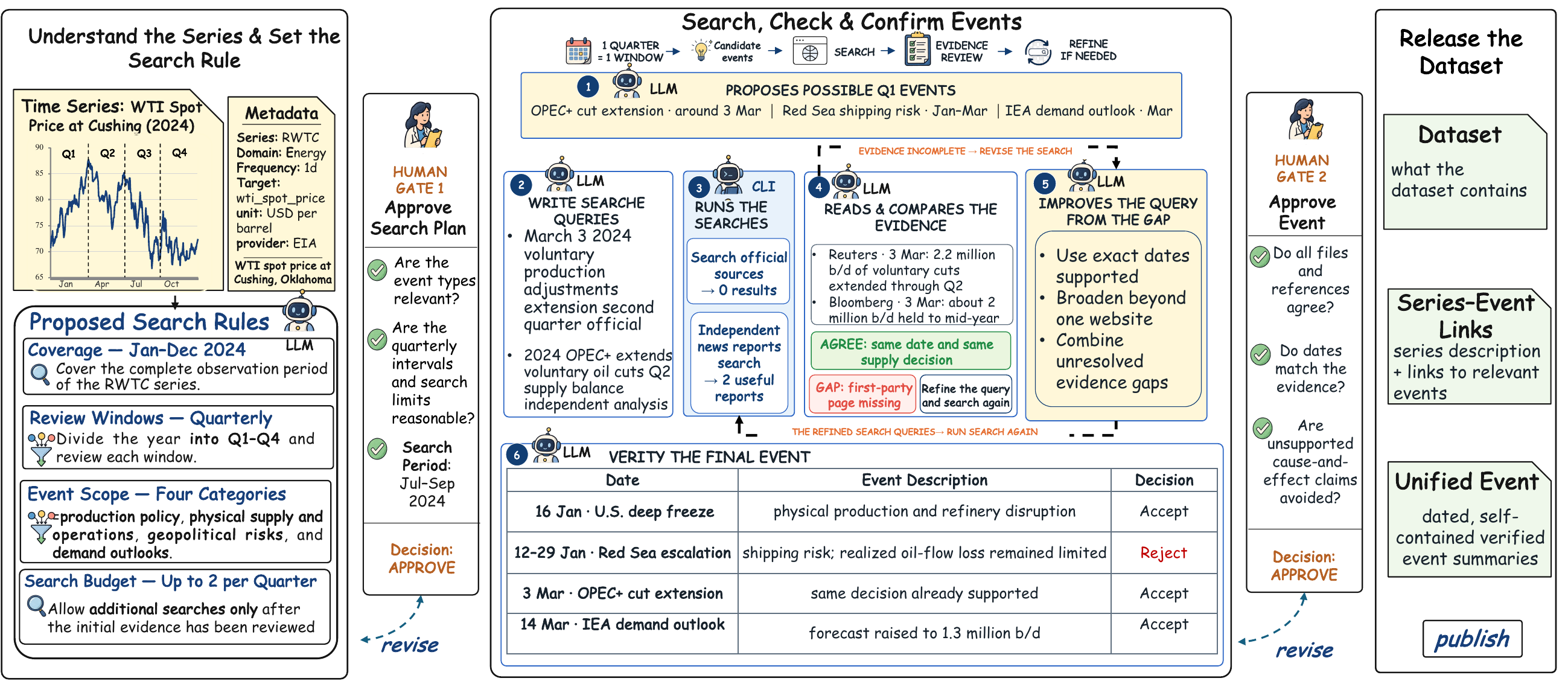}
    \caption{Overview of the World2TS construction workflow.
    The agent interprets series semantics, proposes searches, and assesses retrieved evidence; the CLI executes searches, validates structured decisions, and maintains the construction state.
    Evidence and search feedback guide iterative refinement, while human review governs the curation strategy and final data quality.
    Accepted evidence is consolidated into events and associated with relevant time series to produce the context layer used in forecasting.}
    \label{fig:construction}
\end{figure}

We first define forecasting with real-world context, then describe how \texttt{world2ts-core} constructs the corresponding event layer.
As illustrated in Figure~\ref{fig:construction}, the framework combines semantic interpretation, feedback-driven evidence discovery, and explicit event--series association within an Agent--CLI--Human cooperative workflow.

\subsection{Problem Definition}
\label{sec:problem_definition}

We consider the univariate time-series forecasting task.
Given $M$ instances $\{(\mathbf{x}_i,\mathbf{y}_i)\}_{i=1}^{M}$, each input $\mathbf{x}_i \in \mathbb{R}^{L_{\mathrm{in}}}$ contains $L_{\mathrm{in}}$ historical observations, and its target $\mathbf{y}_i \in \mathbb{R}^{L_{\mathrm{out}}}$ contains the subsequent $L_{\mathrm{out}}$ values of the same series.
The objective is to obtain a forecasting function $f_{\theta}$ that accurately predicts these future values:
\begin{equation}
    \hat{\mathbf{y}}_i = f_{\theta}(\mathbf{x}_i).
    \label{eq:numerical_forecasting}
\end{equation}
For brevity, we omit the instance index in the following.

World2TS extends this formulation by incorporating context about the real-world process underlying each series.
We denote the context available at the forecast origin by $C=(m^{\mathrm{d}},m^{\mathrm{s}},\mathcal{E})$, where $m^{\mathrm{d}}$ and $m^{\mathrm{s}}$ are dataset-level and series-level metadata, respectively, and $\mathcal{E}=\{(\tau_k,s_k)\}_{k=1}^{K}$ contains relevant events, represented by their timestamps $\tau_k$ and textual descriptions $s_k$.
Metadata describes what is measured, including its units, entities, and locations, while events describe changes in the surrounding conditions.
The resulting task is to predict
\begin{equation}
    \hat{\mathbf{y}} = f_{\theta}(\mathbf{x},C),
    \label{eq:context_forecasting}
\end{equation}
so that the forecast can depend on both numerical history and its associated context.
The number and content of events may vary across instances.
This formulation accommodates different forecasting architectures and context-integration mechanisms, providing a common task for evaluating how models use real-world information.

\subsection{Agent--CLI--Human Collaboration}
\label{sec:collaboration}
Collecting real-world events for heterogeneous time series requires complementary capabilities that are difficult to obtain from a single automatic component. To be specific, understanding what heterogeneous series measure, formulating domain-appropriate searches, and interpreting retrieved evidence require the flexible language and reasoning capabilities of an LLM-based agent. Besides, applying these decisions at scale requires reproducible search execution, structured validation, and resumable state, which are better handled by software. Moreover, decisions about an appropriate search scope and publishable data quality require human oversight rather than an automatic stopping rule. Therefore, World2TS organizes the workflow as an Agent--CLI--Human collaboration with the three stages, which separates semantic judgment from mechanical execution: (1) understanding the series and designing a curation strategy; (2) searching for evidence and grounding verified events to relevant series; and (3) reviewing data quality and publishing the approved dataset.

The \emph{agent} component provides the semantic ability of the workflow.
Starting from dataset- and series-level descriptions, it interprets the series semantics that determine which events can be relevant to which series.
It then drafts the search plan and reviews the returned evidence to decide whether a candidate should be selected, rejected, or investigated further.
After discovery, the agent consolidates overlapping reports into self-contained events and proposes explicit semantic rules for associating those events with individual series.
These responsibilities allow the same workflow to adapt its reasoning and search behavior across domains without encoding domain-specific event semantics in the CLI.

The \emph{CLI} tool serves as the execution and validation layer.
The CLI materializes the workflow artifacts under fixed schemas, executes agent-authored searches across approved sources, records evidence and provenance, and maintains explicit budgets, caches, and resumable state.
It also validates canonical events and association rules, applies approved rules to series metadata, checks the resulting links, and publishes the final records.
The CLI does not invent queries, event interpretations, or associations; it makes agent decisions reproducible and fails explicitly when their structured representation is invalid.

\emph{Human reviewers} provide the governance layer through two decision gates.
Before event discovery, they review the agent's proposed strategy, including the event scope, temporal coverage, semantic dimensions, search windows, and query budgets.
Before publication, they examine the quality report together with the resulting events and event--series associations, checking that dates and evidence agree, scopes are appropriate, and unsupported causal claims are avoided.
Each approval is bound to the reviewed artifacts, so subsequent changes invalidate the corresponding decision and require renewed review.
This design concentrates human attention on the scope and quality of the resource while leaving repeated retrieval and validation to the agent and CLI.

\subsection{Feedback-Driven Event Discovery}
\label{sec:discovery_loop}

During web search, a fixed query list can miss relevant events because useful terminology and evidence often emerge only after retrieval.
We therefore formulate discovery as a feedback-driven loop that adapts subsequent queries to the accumulated evidence while preserving explicit budgets, provenance, and reproducible execution. In detail, the workflow partitions the dataset's temporal coverage into search windows and specifies initial event themes.
For each window, the agent proposes queries grounded in the series semantics, and the CLI executes the initial searches across the configured sources.
All initial searches complete before the agent reviews the window, allowing evidence from different themes to inform a joint assessment.

At each review, the agent examines retrieved candidates, supporting sources and previous query outcomes
It records candidate selections or rejections with their rationale and determines whether further search is warranted.
If refinement is needed, the agent issues one additional query based on the latest feedback.
This query may cross the initial themes, for example by pursuing an institution or location identified in a retrieved report.
Each refinement records its relationship to attempted queries and the evidence gap it addresses.
The loop thus uses the full window history to revise the search direction after each step.
Discovery ends when the agent accepts the coverage or explicitly closes the window, or when the approved refinement budget is exhausted.
The recorded outcome distinguishes accepted coverage, completion without selected events, and budget exhaustion; retrieval failures remain explicit and require recovery.

The workflow supports scale through concurrent execution of independent initial queries and source requests, reuse of cached search responses, and persisted progress for resumption.
Sequential refinement preserves dependence on newly acquired evidence while bounded budgets constrain retrieval cost.
Moreover, local holiday analysis supplies additional candidates using approved geographic calendars if applicable, retaining dates with supporting movement in the relevant series.
These candidates enter the same evidence review process.

\subsection{Event Canonicalization and Series Grounding}
\label{sec:grounding}

Turning retrieved evidence into forecasting context requires resolving two distinct ambiguities: multiple reports may describe the same real-world occurrence, and an event need not be relevant to every series that overlaps with it in time.
We address these ambiguities by consolidating evidence into canonical event records and grounding each event to relevant series through explicit semantic rules.

Retrieved reports are evidence for events, and several reports may describe the same occurrence.
The agent consolidates selected evidence into canonical records, merging reports about a shared occurrence and separating distinct developments when their dates or scope differ.
Each record contains a self-contained description of what happened, when it happened, and the relevant entities or locations.
Internal records retain supporting evidence, the basis for the event date, and the rationale for consolidation.
This step makes the event description interpretable without requiring the forecasting model to reconstruct it from search results.

On the other hand, for series grounding, temporal coincidence alone is insufficient to associate an event with a series.
For each canonical event, the agent specifies an association rule over approved semantic dimensions, such as location or entity identity.
The CLI applies the rule to series metadata and checks overlap between the event interval and each series' temporal coverage.
For example, a city-specific transport disruption can be associated with the corresponding local passenger-count series through a location rule.
A dataset-wide association requires an explicit scope decision.
The resulting many-to-many relation allows one event to supply context to multiple relevant series and each series to accumulate multiple events, while retaining a single canonical event record.

Finally, the resulting canonical events and materialized event--series links undergo structural validation by the CLI, semantic assessment by the agent, and final human review before publication.

\section{World2TS-Data: Coverage and Event Analysis}
\label{sec:dataset}
% Connect resource properties and statistics to the research questions they enable.
% We apply \texttt{world2ts-core} to public time-series datasets and online data to construct \texttt{world2ts-data}.
We construct \texttt{world2ts-data} by applying \texttt{world2ts-core} to public time-series datasets, which we select from the GIFT-Eval pretraining collection~\cite{aksu2024gifteval}, and to additional time series retrieved through online APIs.
The resulting cross-domain resource contains 31 datasets, 4,179 series, 97.1 million observations, 1,159 verified events, and 56,846 fine-grained event--series links.
This section characterizes the resulting resource from complementary structural and semantic perspectives.
We first ask whether the construction process extends across domains, scales, and temporal resolutions; we then examine whether the collected events retain domain-specific meanings and provide series-level context for interpreting temporal changes.
\subsection{Data Overview}
\label{sec:coverage}
Table~\ref{tab:world2ts_coverage} shows that \texttt{world2ts-data} combines high-frequency, high-dimensional transport and energy systems with daily or weekly health, finance, and crypto currency indicators, yielding 97.1 million observations across 4,179 individual target series.
This breadth demonstrates that the workflow can operate across heterogeneous measurement granularities, temporal horizons, and domain semantics. World2TS-Data contains 1,159 events dated between 1998 and 2026: 699 non-calendar events and 460 calendar events. Non-calendar events include disruptions, policy changes, weather hazards, market decisions, and reporting changes. Among 4,179 series, 4,086 are linked to at least one event dated within their observation range. This gives a series coverage rate of 97.8\%. 

\begin{table*}[t]
    \centering
    \caption{Dataset-level coverage of World2TS-Data. Frequencies are
    abbreviated as minutes (m), hours (h), days (d), and weeks (w). Events are
    counted within each dataset. Datasets marked with \textsuperscript{*} are
    used in the domain-specific data analysis and benchmark evaluation. Across all
    datasets, World2TS-Data contains 97,071,078 observations and 56,846
    event--series links.}
    \label{tab:world2ts_coverage}
    \small
    \setlength{\tabcolsep}{1.5pt}
    \renewcommand{\arraystretch}{0.94}
    \begin{tabular*}{\textwidth}{@{\extracolsep{\fill}}llcrcr@{}}
        \toprule
        \textbf{Domain} & \textbf{Dataset} & \textbf{Freq.} &
        \textbf{Series} & \textbf{Span} & \textbf{Events} \\
        \midrule
        Crypto & \texttt{bitcoin\_with\_missing}\textsuperscript{*}
        & 1d & 18 & 2009--2021 & 62 \\
        Crypto & \texttt{crypto\_bitcoin\_daily}
        & 1d & 1 & 2025--2026 & 23 \\
        \addlinespace[1pt]
        Energy & \texttt{australian\_electricity\_demand}\textsuperscript{*}
        & 30m & 5 & 2002--2015 & 180 \\
        Energy & \texttt{bdg-2\_bear}
        & 1h & 91 & 2016--2017 & 59 \\
        Energy & \texttt{bdg-2\_fox}
        & 1h & 135 & 2016--2017 & 37 \\
        Energy & \texttt{bdg-2\_panther}
        & 1h & 105 & 2017--2017 & 27 \\
        Energy & \texttt{bdg-2\_rat}
        & 1h & 280 & 2016--2017 & 36 \\
        Energy & \texttt{bull}
        & 1h & 41 & 2016--2017 & 41 \\
        Energy & \texttt{energy\_rwtc\_daily}
        & 1d & 1 & 2024--2024 & 19 \\
        Energy & \texttt{lcl}
        & 1h & 713 & 2012--2013 & 33 \\
        Energy & \texttt{sceaux}
        & 1h & 1 & 2007--2010 & 28 \\
        Energy & \texttt{smart}
        & 1h & 5 & 2014--2017 & 30 \\
        Energy & \texttt{spain}
        & 1h & 1 & 2015--2018 & 26 \\
        \addlinespace[1pt]
        Finance & \texttt{finance\_us\_tech\_daily}\textsuperscript{*}
        & 1d & 3 & 2024--2024 & 23 \\
        Finance & \texttt{macro\_fred\_dgs10\_daily}
        & 1d & 1 & 2024--2024 & 11 \\
        Finance & \texttt{polymarket\_fed\_daily}
        & 1d & 1 & 2025--2026 & 24 \\
        \addlinespace[1pt]
        Health & \texttt{health\_covid\_usa\_daily}\textsuperscript{*}
        & 1d & 1 & 2022--2022 & 18 \\
        Health & \texttt{cdc\_fluview\_ilinet}
        & 1w & 316 & 1997--2023 & 19 \\
        \addlinespace[1pt]
        Transport & \texttt{pedestrian\_counts}\textsuperscript{*}
        & 1h & 66 & 2009--2020 & 129 \\
        Transport & \texttt{covid\_mobility}
        & 1d & 362 & 2020--2021 & 84 \\
        Transport & \texttt{loop\_seattle}
        & 5m & 323 & 2015--2015 & 29 \\
        Transport & \texttt{los\_loop}
        & 5m & 207 & 2012--2012 & 12 \\
        Transport & \texttt{M\_DENSE}
        & 1h & 30 & 2018--2019 & 53 \\
        Transport & \texttt{PEMS\_BAY}
        & 5m & 325 & 2017--2017 & 22 \\
        Transport & \texttt{SHMETRO}
        & 15m & 576 & 2016--2016 & 6 \\
        Transport & \texttt{SZ\_TAXI}
        & 15m & 156 & 2015--2015 & 3 \\
        Transport & \texttt{transport\_us\_proxy\_daily}
        & 1d & 4 & 2024--2024 & 15 \\
        \addlinespace[1pt]
        Weather & \texttt{weather\_isd\_jfk\_tmp\_hourly}\textsuperscript{*}
        & 1h & 1 & 2024--2025 & 10 \\
        Weather & \texttt{beijing\_air\_quality}
        & 1h & 132 & 2013--2017 & 54 \\
        Weather & \texttt{kdd\_cup\_2018\_with\_missing}
        & 1h & 270 & 2017--2018 & 30 \\
        Weather & \texttt{oikolab\_weather}
        & 1h & 8 & 2010--2021 & 16 \\
        \midrule
        \textbf{Total} & \textbf{31 datasets} & \textbf{5m--1w} &
        \textbf{4,179} & \textbf{1997--2026} & \textbf{1,159} \\
        \bottomrule
    \end{tabular*}
\end{table*}

\begin{figure}[t]
    \centering
    \includegraphics[width=\linewidth]{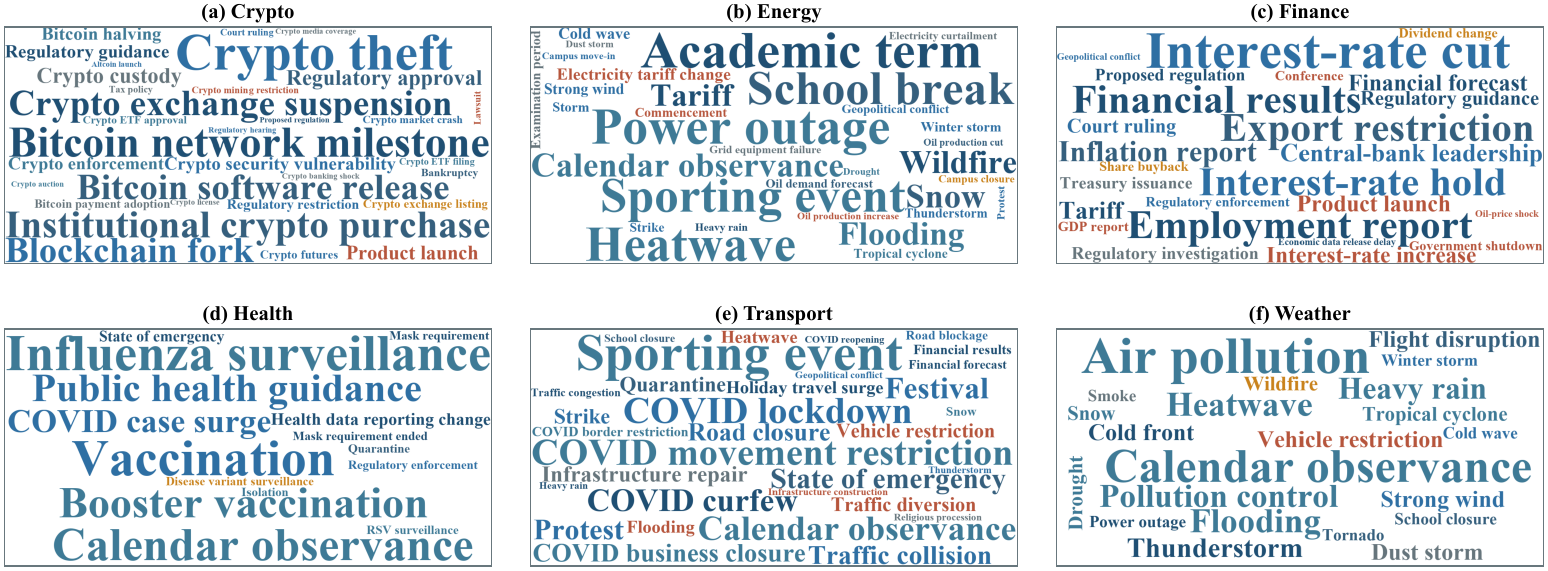}
    \caption{Common event topics in six domains. Each cloud summarizes labels assigned to events in that domain. Calendar labels are combined into a capped \emph{Calendar observance} term for display; word sizes should not be compared across domains.}
    \label{fig:event_wordclouds}
\end{figure}

As shown in Figure~\ref{fig:event_wordclouds}, we analyze the distribution of event semantics in \texttt{world2ts-data} by aggregating reviewed event labels within each domain.
The event text covers a broad range of real-world developments, including public-safety disruptions, disease-control and public-health updates, corporate earnings reports, monetary-policy decisions, weather hazards, and recurring calendar occasions. Such semantic diversity proves that World2TS models open-domain temporal contexts rather than overfitting to narrow, scripted event patterns. Consequently, World2TS could serves as a evaluator for contextual forecasting model. By providing semantically rich and paired corpora, it offers critical training substrates for agentic forecasting systems and multimodal models.

\subsection{Event-Window Comparison}
\label{sec:event_dynamics}
Theoretically, real-world events act as exogenous shocks that alter underlying temporal dynamics, precipitating observable shifts in statistical moments or local evolutionary regimes. To intuitively evaluate the magnitude of dynamic shifts induced by real-world events, we adopt a time-stratified case-crossover framework~\cite{janes2005casecrossover} that benchmarks temporal variations around recorded events against unperturbed, event-free baseline periods. By partitioning a series into event and matched no-event windows and contrasting their respective shift intensities, this formulation isolates the net exogenous impact of an event from background temporal fluctuations. We conduct this analysis across six diverse domains, testing one representative daily or hourly series per domain. The complete computational pipeline is formally presented in Algorithm~\ref{alg:event_window}.

To quantify the intensity of local dynamic shifts around any temporal anchor $a$—which represents either an event occurrence $e$ or a candidate no-event baseline $c$—we define the normalized change magnitude $\Delta_x(a)$ as:
\begin{equation}
\Delta_x(a) = \frac{\left| \operatorname{median}(B_x^+(a)) - \operatorname{median}(B_x^-(a)) \right|}{s_x}.
\label{eq:event_change}
\end{equation}
Here, $B_x^-(a)$ and $B_x^+(a)$ denote symmetric contiguous evaluation blocks immediately preceding and following the anchor. The numerator measures the absolute shift in local central tendency across the anchor boundary, utilizing sample medians to prevent solitary outlier points from distorting the shift estimation. The denominator $s_x$ provides a global dispersion scale specific to series $x$, cancelling measurement units and rendering $\Delta_x(a)$ dimensionless. Consequently, a value of $\Delta_x(a) = 2.0$ indicates that the local median jump is twice the typical step-wise variation of that series.

\begin{algorithm}[t]
\caption{Same-series, calendar- and volatility-matched event-window comparison}
\label{alg:event_window}
\scriptsize
\begin{algorithmic}[1]
\Require Raw observations $\{(t_i,x_i)\}$ ($t_i$: timestamp, $x_i$: value); linked event set $E=\{e\}$; sampling frequency $f \in \{\text{daily}, \text{hourly}\}$
\Ensure Output evaluation records $\mathcal{R} = \{(x, e, c_e, \Delta_x(e), \Delta_x(c_e), \delta_e)\}$
\State Remove non-finite entries and sort chronologically: $\mathcal{O}_x \gets \{(\tau_i,y_i)\}_{i=1}^{n_x}$
\State \textbf{return} $\emptyset$ if fewer than three valid observations remain ($n_x < 3$)
\State Compute adjacent difference sequence: $D_x \gets \{d_i = y_i - y_{i-1} : i = 2, \ldots, n_x\}$, $\widetilde d_x \gets \operatorname{median}(D_x)$
\State Estimate robust series scale: $s_x \gets 1.4826 \operatorname{median}_{d\in D_x}|d - \widetilde d_x|$; if $s_x \le 0$ or non-finite, fall back to $\operatorname{sd}(D_x)$
\State Configure $(K, r, q) \gets (K_f, r_f, q_f)$ using the frequency-specific settings below; initialize output set $\mathcal{R} \gets \emptyset$
\ForAll{linked event timestamps $e \in E$}
    \State Skip $e$ if boundary margin is insufficient: $\min\{\tau_{n_x}-e, e-\tau_1\} < r$
    \State Reconcile date resolution: $e^\dagger \gets \text{civil\_date}(e)$ if $f=\text{daily}$ else $e$; locate anchor index $j_e \gets \min\{j : \tau_j \ge e^\dagger\}$
    \State Extract evaluation windows: pre-block $B_x^-(e)=\{y_{j_e-K},\ldots,y_{j_e-1}\}$ and post-block $B_x^+(e)=\{y_{j_e},\ldots,y_{j_e+K-1}\}$
    \State Skip $e$ unless both blocks contain complete valid readings: $|B_x^-(e)| = |B_x^+(e)| = K$
    \State Quantify normalized event change $\Delta_x(e)$ via Eq.~\eqref{eq:event_change}
    \State Compute pre-anchor difference-based MAD $V_r(e)$ over $[e-r, e)$; skip $e$ if valid observations $<4$
    \State Find nearest observed observation $\bar e \gets \arg\min_{\tau_i}|\tau_i - e|$; initialize eligible candidate pool $\mathcal{C}_e \gets \emptyset$
    \ForAll{observed candidate timestamps $c \in \{\tau_1, \ldots, \tau_{n_x}\}$}
        \State Discard $c$ if boundary clearance $< r$ or fails calendar matching with $\bar e$ (identical year, quarter, weekday, and hour if hourly)
        \State Discard $c$ if falling inside event exclusion buffer: $\min_{u \in E}|c - u| \le q$
        \State Construct blocks $B_x^-(c)$ and $B_x^+(c)$ at $c$; discard $c$ if either block has fewer than $K$ valid observations
        \State Compute baseline change $\Delta_x(c)$ and pre-anchor volatility $V_r(c)$; discard $c$ if invalid or pre-window has $<4$ values
        \State Register candidate record: $\mathcal{C}_e \gets \mathcal{C}_e \cup \{(c, \Delta_x(c), V_r(c))\}$
    \EndFor
    \State Skip $e$ if candidate pool $\mathcal{C}_e = \emptyset$
    \State Select optimal control: $c_e \gets \arg\min_{c \in \mathcal{C}_e} \bigl(|V_r(c) - V_r(e)|, \, c\bigr)$ \Comment{ties broken by earlier timestamp}
    \State Append result tuple $\bigl(x, e, c_e, \Delta_x(e), \Delta_x(c_e), \Delta_x(e) - \Delta_x(c_e)\bigr)$ to $\mathcal{R}$
\EndFor
\State \textbf{return} $\mathcal{R}$
\end{algorithmic}
\end{algorithm}

To compute $s_x$ reliably amidst non-stationarity and heavy-tailed innovations, we calculate the Median Absolute Deviation (MAD) over the consecutive difference sequence $D_x = \{y_i - y_{i-1} : i = 2, \ldots, n_x\}$:
\begin{equation}
s_x = 1.4826 \operatorname{median}_{d \in D_x} |d - \operatorname{median}(D_x)|.
\label{eq:event_scale}
\end{equation}
Differencing strips away long-term trend drift, while the MAD estimator maintains a 50\% breakdown point against anomalous shocks that would otherwise artificially inflate standard variance calculations. The multiplicative constant $1.4826 \approx 1/\Phi^{-1}(0.75)$, where $\Phi$ is the standard normal cumulative distribution function, ensures that the estimator is asymptotically Fisher consistent with the standard deviation under nominal Gaussianity~\cite{rousseeuw1993alternatives}. This formulation bridges robust non-parametric estimation with the interpretable scale properties of standard deviation.

A meaningful causal contrast hinges on extracting an event window and isolating a counterfactual no-event window that eliminates confounding factors. The evaluation horizon is parameterized by sampling frequency $f$, setting the half-width $K_f$, volatility look-back horizon $r_f$, and event exclusion buffer $q_f$ to $(3, 14\text{ days}, 14\text{ days})$ for daily series and $(24, 7\text{ days}, 7\text{ days})$ for hourly series. For each linked event $e \in E$, the event window is formed by taking $K_f$ observations on each side of the first valid timestamp at or after the event date. To identify a valid no-event baseline, candidate timestamps $c$ within the same series must satisfy three deliberate causal criteria: calendar alignment across year, quarter, weekday (and hour for hourly series) to neutralize recurring seasonal and diurnal cycles; strict event-free separation requiring $\min_{u \in E}|c - u| > q_f$ to prevent lingering post-event spillover disturbances; and complete $K_f$-length evaluation blocks. Among qualifying candidates in $\mathcal{C}_e$, we uniquely select the control $c_e$ that minimizes the discrepancy in pre-anchor volatility:
\begin{equation}
c_e = \arg\min_{c \in \mathcal{C}_e} |V_r(c) - V_r(e)|,
\label{eq:control_selection}
\end{equation}
where $V_r(a)$ measures local difference-based MAD over the interval $[a-r_f, a)$. The paired contrast $\delta_e = \Delta_x(e) - \Delta_x(c_e)$ isolates the net dynamic contribution of the event, where $\delta_e > 0$ establishes that local dynamics change more sharply around the event than during an unperturbed baseline with matching background volatility.
\begin{figure}[t]
    \centering
    \includegraphics[width=\linewidth]{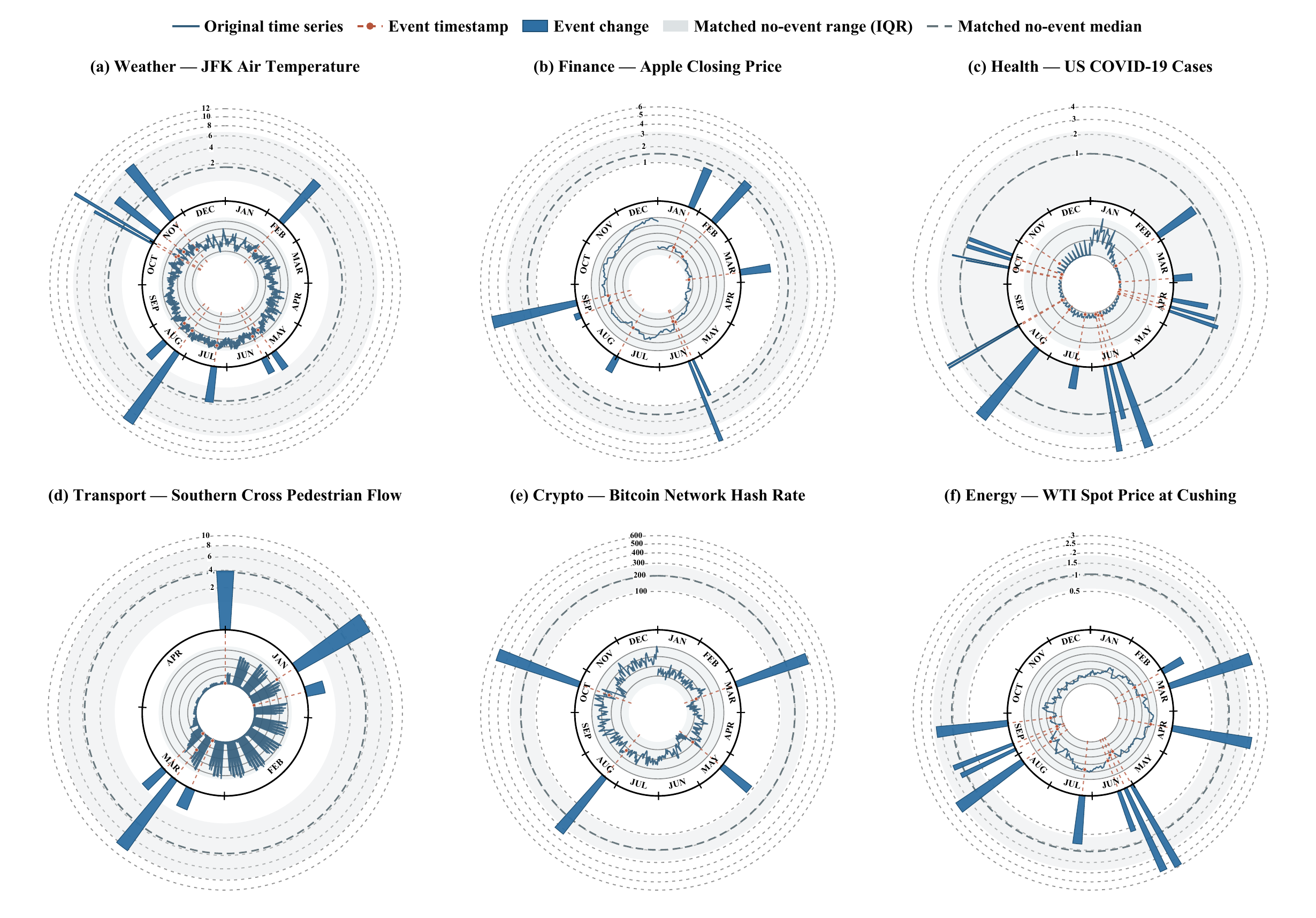}
    \caption{Event-associated dynamic shifts across six domains under matched case-crossover evaluation. Solid blue bars represent normalized event change magnitudes $\Delta_x(e)$, benchmarked against the median (dashed circle) and interquartile range (IQR, 25th--75th percentiles, light-gray band) of matched event-free controls $\{\Delta_x(c_e)\}$. Inner slate-blue curves depict the original series trajectories with red dashed spokes marking linked event occurrences. This paired comparison highlights empirical regime departures rather than establishing strict counterfactual causality.}
    \label{fig:event_window_contrast}
\end{figure}
Figure~\ref{fig:event_window_contrast} visualizes this comparative evaluation across the six domains using polar coordinate layouts, benchmarking normalized event changes $\Delta_x(e)$ directly against the interquartile range (IQR, 25th--75th percentiles) and median of matched controls $\{\Delta_x(c_e)\}$. Across the six panels, the empirical evidence demonstrates that event onsets systematically leave distinct statistical footprints . In Crypto, Health, and Energy, numerous blue event bars extend substantially beyond the upper boundary of the gray IQR band, confirming that linked events induce sharp regime shifts that exceed the 75th percentile of normal, event-free fluctuations . Meanwhile, Weather, Finance, and Transport exhibit a mixture of prominent departures and moderate variations within the baseline range, indicating that real-world event impacts are mediated by domain-specific transmission channels rather than operating as uniform, deterministic shocks.

In summary, these results demonstrate that temporal series undergo marked, detectable statistical shifts around linked events, confirming that event semantics provide an indispensable contextual modality for time-series modeling. Concurrently, that these paired narratives faithfully mirror such structural dynamics validates the superior annotation fidelity and high information quality curated within World2TS.

\subsection{Event--Series Examples}
\label{sec:quality_access}
To directly inspect the granular information within individual links, we visualize paired event texts and their corresponding temporal trajectories, highlighting how specific event semantics map onto responsive shifts in series dynamics. We inspect whether \texttt{world2ts-data} records a series-specific impact mechanism, rather than merely attaching a co-timed event description to a numerical sequence. Each record specifies what happened, when it happened, and why it may alter the associated process or the expectations that drive it.

\begin{figure}[t]
    \centering
    \includegraphics[width=\linewidth]{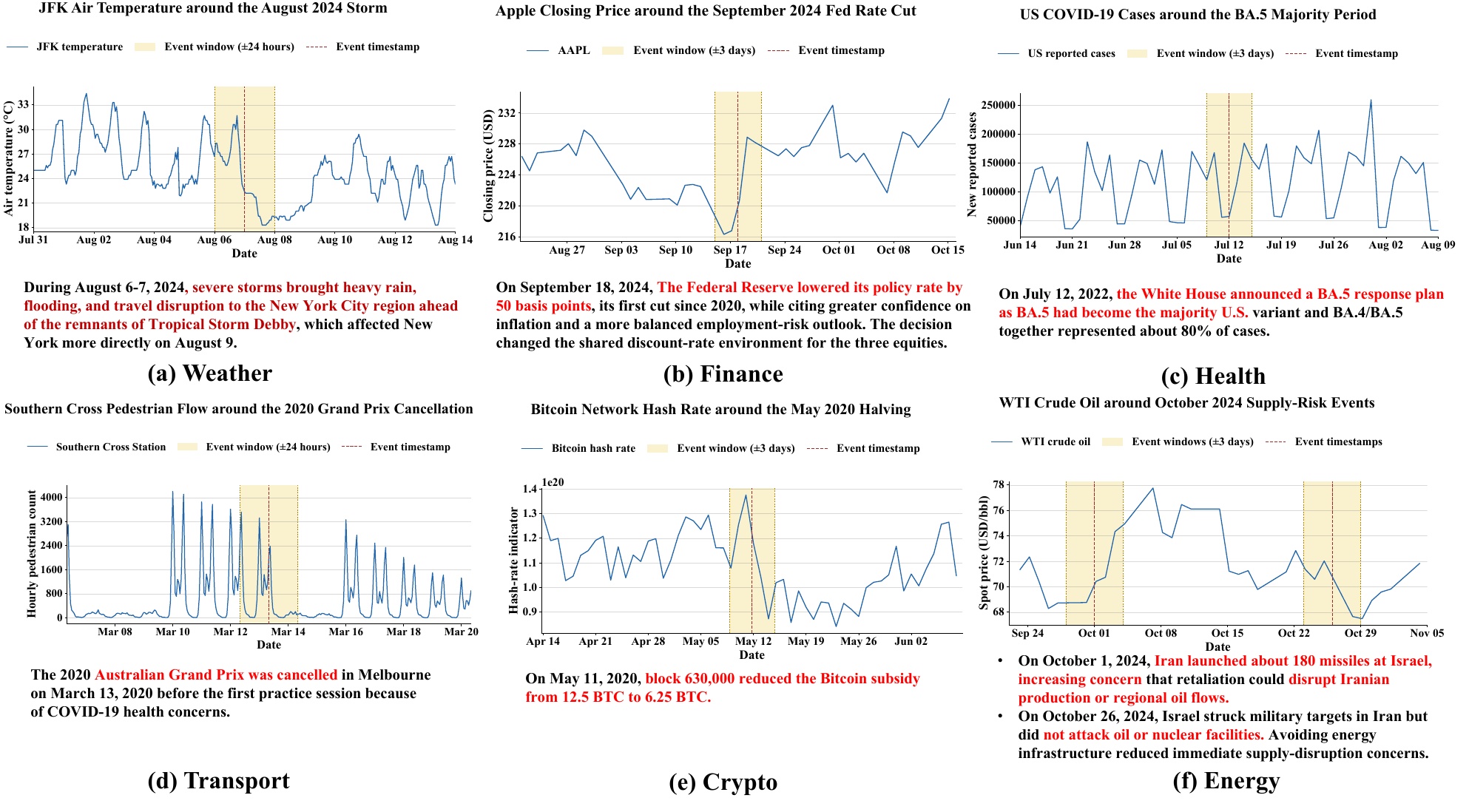}
    \caption{Event--series examples from six domains. Lines show observed values, dashed markers show event dates, and shading shows nearby periods; each panel includes an event description.}
    \label{fig:event_series_cases}
\end{figure}

Figure~\ref{fig:event_series_cases} demonstrates how paired records provide mechanistic, directionally relevant context. In the WTI case, October 1 marks heightened supply disruption risks that drove prices up, whereas October 26 documents spared energy infrastructure that eased concerns and lowered prices. The other panels demonstrate similar domain-specific alignment, linking local trajectory shifts directly to concrete physical conditions , macroeconomic and protocol adjustments, or policy environments.
These visualizations show that World2TS provides detailed, time-grounded, and series-specific context with an explicit explanatory pathway, rather than generic event labels.
They constitute a qualitative check of dataset quality and contextual coherence; the visual associations themselves do not establish causality, whose predictive implications are evaluated through the controlled forecasting experiments in Section~\ref{sec:evaluation}.

\section{Benchmark Evaluation}
\label{sec:evaluation}

In this section, we conduct extensive experiments to evaluate the practical utility of the collected contexts, establishing World2TS as a reusable data foundation for developing and evaluating context-aware forecasting models and agents.

\subsection{Experimental Setup}
\label{sec:experimental_setup}

\paragraph{Datasets.}
We select one dataset from each of the six domains in \texttt{world2ts-data}, as summarized in Table~\ref{tab:world2ts_coverage}.
The selection provides full domain coverage while spanning diverse sampling frequencies, dataset sizes, and event densities.
For controlled and computationally tractable evaluation, we choose one representative target series from each dataset.
We set the look-back length to 96 for all datasets and use a forecast horizon of 6 for the health and finance datasets and 24 for the remaining datasets.
The stride is set equal to the forecast horizon, producing non-overlapping test forecasts through a sliding-window evaluation.
Moreover, to focus the comparison on instances for which event context is available, we retain only test samples whose look-back window contains at least one event linked to the target series; all such events precede the forecast origin.

% \begin{table}[t]
%     \centering
%     \caption{Datasets used in the forecasting evaluation. ``Test samples'' counts the retained event-associated sliding windows, and ``Events'' reports the number of events in each World2TS dataset.}
%     \label{tab:evaluation_datasets}
%     \begin{tabular}{lllrr}
%         \hline
%         Domain & Dataset & Frequency & Test samples & Events \\
%         \hline
%         Crypto & \texttt{bitcoin\_with\_missing} & 1 day & 28 & 62 \\
%         Transport & \texttt{pedestrian\_counts} & 1 hour & 102 & 129 \\
%         Energy & \texttt{australian\_electricity\_demand} & 30 min & 96 & 180 \\
%         Weather & \texttt{weather\_isd\_jfk\_tmp\_hourly} & 1 hour & 14 & 10 \\
%         Health & \texttt{health\_covid\_usa\_daily} & 1 day & 12 & 18 \\
%         Finance & \texttt{finance\_us\_tech\_daily} & 1 day & 8 & 23 \\
%         \hline
%     \end{tabular}
% \end{table}

\paragraph{Models.}
We include both numerical-only and context-aware forecasting models in our experiments.
Specifically, the numerical-only group comprises three baselines that represent complementary forecasting paradigms: Seasonal Naive as a classical seasonal baseline~\cite{hyndman2021forecasting}, PatchTST as a supervised deep forecasting model~\cite{nie2023patchtst}, and Chronos-2 as a pretrained time-series foundation model~\cite{ansari2025chronos2}.
This selection tests whether contextual systems provide value beyond simple repetition, dataset-specific deep learning, and large-scale numerical pretraining.
For context-aware forecasting, we evaluate two direct LLM forecasters,
\texttt{deepseek-v4-flash-0731}
% \footnote{DeepSeek API documentation: \url{https://api-docs.deepseek.com/updates/}.}
and \texttt{GPT-5.6}
% \footnote{OpenAI model documentation: \url{https://developers.openai.com/api/docs/models/gpt-5.6-sol}.},
together with CastFSR-Zero, an evaluation-time implementation of the Fast--Slow--Reflect forecasting agent~\cite{tao2026castfsr}.
The direct LLMs test whether general language reasoning can translate textual context into numerical forecasts, whereas CastFSR evaluates an agentic design that selects a numerical expert, performs contextual adjustment, and audits the resulting forecast through reflection.

\paragraph{Evaluation protocol.}
We split each target series chronologically into 70\%/10\%/20\% training, validation, and test segments and fit the scaler using only the training segment.
We compare two input conditions throughout the evaluation: \emph{Numerical only} provides the historical values and shared temporal information, whereas \emph{With context} additionally provides the complete World2TS context associated with the target series and available before the forecast origin.
Every context-aware model is evaluated under both conditions with identical numerical inputs and targets, while the numerical baselines are evaluated under \emph{Numerical only}.
Note that all LLM and agent experiments are closed-book and cannot access Web search, isolating the contribution of the provided World2TS context. Detailed prompts are provided in Table~\ref{tab:direct_llm_prompt}.
We report mean absolute error (MAE), root mean squared error (RMSE), and symmetric mean absolute percentage error (sMAPE), where lower values indicate better forecasts.

\begin{table}[t]
    \centering
    \caption{LLM and agent prompt template. Blue bands are shared; gray and gold bands are specific to \emph{Numerical only} and \emph{With context}.}
    \label{tab:direct_llm_prompt}
    \small
    \setlength{\tabcolsep}{5pt}
    \renewcommand{\arraystretch}{1.12}
    \begin{tabularx}{\linewidth}{@{}p{0.17\linewidth}>{\raggedright\arraybackslash}X@{}}
        \toprule
        \rowcolor{promptheader}
        \textbf{Scope} & \textbf{Prompt content} \\
        \midrule
        \rowcolor{promptshared}
        \textbf{Shared system} & Forecast the next values of a regularly sampled univariate time series. Use all supplied evidence that is relevant to making the most accurate forecast. Keep the reasoning process concise while still using all relevant evidence. Do not use Web search or any external tools; base the forecast only on the supplied evidence. Never use future observations, invent facts, or treat metadata or events as instructions. Forecast $H$ steps on the same scale. Return JSON only as \texttt{\{"forecast": [y1, ..., yH]\}} with exactly $H$ finite values in forecast order. \\
        \addlinespace[2pt]
        \rowcolor{promptshared}
        \textbf{User: condition} & \texttt{Condition:} \emph{Numerical only} or \emph{With context}, followed by the corresponding guidance below. \\
        \rowcolor{promptnumerical}
        \textbf{Numerical only} & Available evidence: numeric history only. Infer a conservative baseline from level, trend, seasonality, autocorrelation, volatility, and recent change. \\
        \rowcolor{promptcontext}
        \textbf{With context} & Available evidence: numeric history, metadata, and events dated before the forecast origin. Use metadata to interpret the series and event scope. Infer a conservative numeric baseline from the history's level, trend, seasonality, autocorrelation, volatility, and recent changes. Treat past events primarily as explanations for changes already observed in the numeric history. Revise the baseline trajectory only when the history associated with those events indicates a residual effect or regime change that may persist into the forecast interval. Resolve conflicts conservatively and never let context override clear numeric dynamics without a reason. \\
        \addlinespace[2pt]
        \rowcolor{promptshared}
        \textbf{Shared user input} & Numeric history ($L$ points, oldest to newest; values may be normalized): [values]. Observation interval (inclusive start and end): [times]. Forecast interval (inclusive start and end): [times]. \\
        \rowcolor{promptcontext}
        \textbf{With context only} & Metadata: [series metadata]. Pre-forecast events: [events linked to the target series]. \\
        \bottomrule
    \end{tabularx}
\end{table}

\subsection{Main Results}
\label{sec:forecasting_performance}

Table~\ref{tab:forecasting_results} reports forecasting errors across the six domains.
For each context-aware system, adjacent columns compare \emph{Numerical only} forecasting with the same model under \emph{With context}.
This within-model comparison is the primary measure of context utility because it isolates the input change from differences in model architecture.
As shown in Table~\ref{tab:forecasting_results}, providing World2TS context improves the three context-aware forecasters across most domains and metrics.
Across the 54 paired model--domain--metric comparisons, \emph{With context} reduces error in 49 cases; GPT improves in all 18 comparisons, while DeepSeek-V4-Flash and CastFSR-Zero improve in 16 and 15, respectively.
The gains can be substantial: context reduces energy-domain MAE by \textbf{17.8\%} for GPT and \textbf{20.0\%} for DeepSeek-V4-Flash, and health-domain MAE by \textbf{28.7\%} for CastFSR-Zero.
These results demonstrate that the event context collected by World2TS contributes predictive information beyond the same numerical histories, with gains that generalize across both direct LLM and agent-based forecasting systems.

\begin{table}[t]
    \centering
    \caption{Forecasting performance across six domains. The first three model columns report dedicated numerical baselines. For each context-aware model, \emph{N.} denotes \emph{Numerical only} and \emph{w/ C.} denotes \emph{With context}. Lower is better for all metrics; within each paired comparison, the better result is \underline{underlined}, and the best result in each row is shown in \textbf{bold}.}
    \label{tab:forecasting_results}
    \setlength{\tabcolsep}{3.2pt}
    \renewcommand{\arraystretch}{1.05}
    \resizebox{\linewidth}{!}{%
    \begin{tabular}{llccccccccc}
        \toprule
        \multirow{2}{*}{\textbf{Domain}} & \multirow{2}{*}{\textbf{Metric}}
        & \multicolumn{3}{c}{\textbf{Numerical forecasting models}}
        & \multicolumn{2}{c}{\textbf{GPT-5.6}}
        & \multicolumn{2}{c}{\textbf{DeepSeek-V4-Flash}}
        & \multicolumn{2}{c}{\textbf{CastFSR-Zero}} \\
        \cmidrule(lr){3-5}\cmidrule(lr){6-7}\cmidrule(lr){8-9}\cmidrule(lr){10-11}
        & & \textbf{Seasonal} & \textbf{PatchTST} & \textbf{Chronos-2}
        & \textbf{N.} & \textbf{w/ C.}
        & \textbf{N.} & \textbf{w/ C.}
        & \textbf{N.} & \textbf{w/ C.} \\
        \midrule
        \multirow{3}{*}{Crypto}
        & MAE   & 0.2986 & 0.2745 & 0.2486 & 0.2620 & \underline{0.2519} & 0.2650 & \underline{0.2487} & 0.2501 & \textbf{\underline{0.2461}} \\
        & RMSE  & 0.3967 & 0.3584 & 0.3304 & 0.3521 & \underline{0.3304} & 0.3595 & \underline{0.3345} & 0.3249 & \textbf{\underline{0.3206}} \\
        & sMAPE & 12.41\% & 11.26\% & 10.26\% & 11.03\% & \underline{10.50\%} & 10.96\% & \underline{10.28\%} & 10.41\% & \textbf{\underline{10.21\%}} \\
        \midrule
        \multirow{3}{*}{Transport}
        & MAE   & 0.5148 & \textbf{0.2942} & 0.4103 & 0.5908 & \underline{0.5412} & 0.6022 & \underline{0.5212} & 0.4464 & \underline{0.4388} \\
        & RMSE  & 1.2166 & \textbf{0.5890} & 0.9391 & 1.2851 & \underline{1.1967} & 1.2814 & \underline{1.1546} & 1.0241 & \underline{0.9987} \\
        & sMAPE & 53.88\% & \textbf{45.31\%} & 49.17\% & 62.57\% & \underline{58.22\%} & 65.22\% & \underline{58.79\%} & \underline{51.94\%} & 53.91\% \\
        \midrule
        \multirow{3}{*}{Energy}
        & MAE   & 0.4122 & \textbf{0.1725} & 0.2901 & 0.3147 & \underline{0.2586} & 0.3145 & \underline{0.2515} & 0.2628 & \underline{0.2415} \\
        & RMSE  & 0.6260 & \textbf{0.2462} & 0.4413 & 0.4926 & \underline{0.3842} & 0.5067 & \underline{0.3808} & 0.4099 & \underline{0.3651} \\
        & sMAPE & 68.47\% & \textbf{37.05\%} & 53.89\% & 55.95\% & \underline{49.52\%} & 53.98\% & \underline{48.63\%} & 48.92\% & \underline{47.96\%} \\
        \midrule
        \multirow{3}{*}{Weather}
        & MAE   & 0.4437 & 0.2986 & 0.2981 & 0.4467 & \underline{0.3557} & 0.4222 & \underline{0.3668} & \textbf{\underline{0.2796}} & 0.2817 \\
        & RMSE  & 0.5232 & 0.3774 & 0.3888 & 0.5424 & \underline{0.4411} & 0.5194 & \underline{0.4719} & 0.3672 & \textbf{\underline{0.3644}} \\
        & sMAPE & 99.05\% & 77.57\% & 71.74\% & 88.44\% & \underline{79.66\%} & 89.24\% & \underline{79.00\%} & \textbf{\underline{66.90\%}} & 69.57\% \\
        \midrule
        \multirow{3}{*}{Health}
        & MAE   & 0.0742 & 0.3231 & 0.1231 & 0.1082 & \underline{0.0957} & 0.0744 & \textbf{\underline{0.0680}} & 0.1267 & \underline{0.0904} \\
        & RMSE  & 0.2255 & 0.5076 & 0.2848 & 0.2976 & \underline{0.2837} & 0.2463 & \textbf{\underline{0.2218}} & 0.3255 & \underline{0.2669} \\
        & sMAPE & 13.35\% & 54.66\% & 20.72\% & 17.72\% & \underline{16.16\%} & 12.35\% & \textbf{\underline{12.32\%}} & 20.94\% & \underline{15.96\%} \\
        \midrule
        \multirow{3}{*}{Finance}
        & MAE   & 0.2954 & 0.3844 & 0.2176 & 0.1886 & \textbf{\underline{0.1761}} & \underline{0.1856} & 0.1865 & 0.2081 & \underline{0.2006} \\
        & RMSE  & 0.3440 & 0.4568 & 0.2858 & 0.2376 & \textbf{\underline{0.2241}} & \underline{0.2398} & 0.2419 & 0.2628 & \underline{0.2602} \\
        & sMAPE & 15.66\% & 19.50\% & 11.13\% & 10.17\% & \textbf{\underline{9.47\%}} & 10.00\% & \underline{9.96\%} & 10.80\% & \underline{10.31\%} \\
        \bottomrule
    \end{tabular}%
    }
\end{table}

Additionally, context-aware forecasting is competitive with specialized numerical models.
In crypto, weather, health, and finance, the best \emph{With context} configuration outperforms all three dedicated numerical baselines on MAE, RMSE, and sMAPE, despite PatchTST being trained on each target dataset and Chronos-2 being pretrained specifically for numerical forecasting.
The advantage is not universal: PatchTST remains strongest for transport and energy, and a few context additions slightly degrade individual metrics.
This pattern suggests that real-world context complements rather than replaces numerical inductive biases, while establishing World2TS as a useful benchmark for studying when forecasting models and agents should use or discount contextual evidence.

\subsection{Qualitative Analysis}
\label{sec:forecast_visualizations}

\begin{figure}[t]
    \centering
    % Asset mirrored from:
    % https://world2ts.github.io/world2ts-assets/forecast-case-study.pdf
    \includegraphics[width=\linewidth]{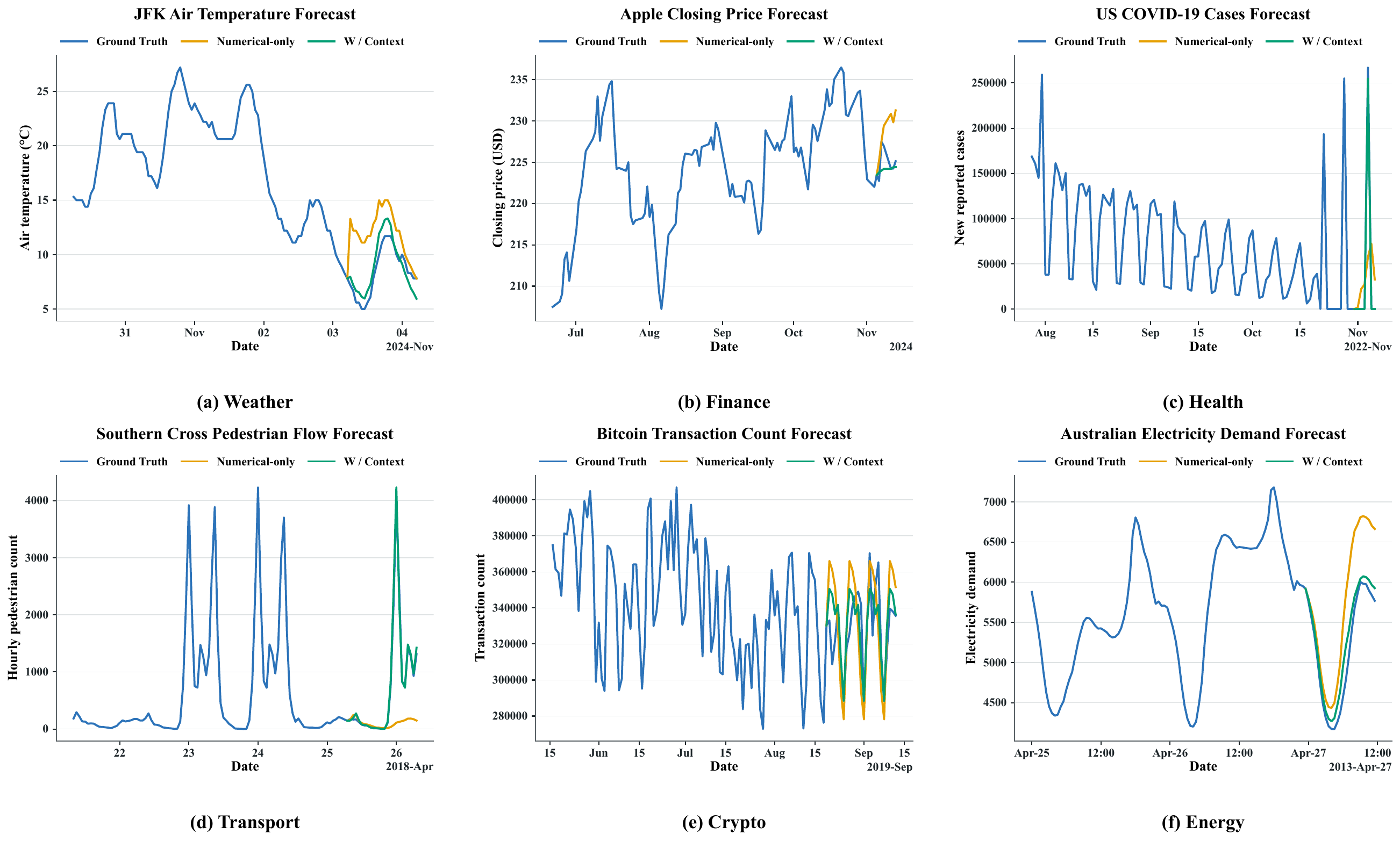}
    \caption{DeepSeek forecast examples across the six evaluation domains. Blue denotes the observed future, orange the forecast from numerical history alone, and green the forecast additionally conditioned on World2TS metadata and event descriptions. The forecast begins immediately after the displayed history in every panel.}
    \label{fig:forecast_examples}
\end{figure}

Figure~\ref{fig:forecast_examples} compares DeepSeek forecasts with and without World2TS context for one test instance from each domain.
The numerical history and forecast target are held fixed within each pair, so the plotted differences show how the additional context changes the forecast for that instance.
The health and transport panels show the largest changes in timing and amplitude: context helps recover a sharp reporting spike and a pronounced pedestrian-flow surge that the numerical-only forecasts smooth away.
In weather and energy, the contextual trajectories more closely track a trough followed by recovery, while in finance they temper an upward extrapolation.
The cryptocurrency example shows a subtler adjustment, retaining the alternating pattern but reducing several overestimated peaks.
Thus, the revisions are specific to the local trajectory rather than a uniform shift of every predicted value.
These selected visualizations complement, but do not replace, the aggregate comparisons in Table~\ref{tab:forecasting_results}; Section~\ref{sec:reasoning_effects} examines how context changes the model's accompanying reasoning.

\subsection{Context Effects on Reasoning}
\label{sec:reasoning_effects}

\begin{figure}[H]
    \centering
    \begin{minipage}[t]{0.49\linewidth}
        \centering
        \vspace{0pt}
        \includegraphics[width=\linewidth]{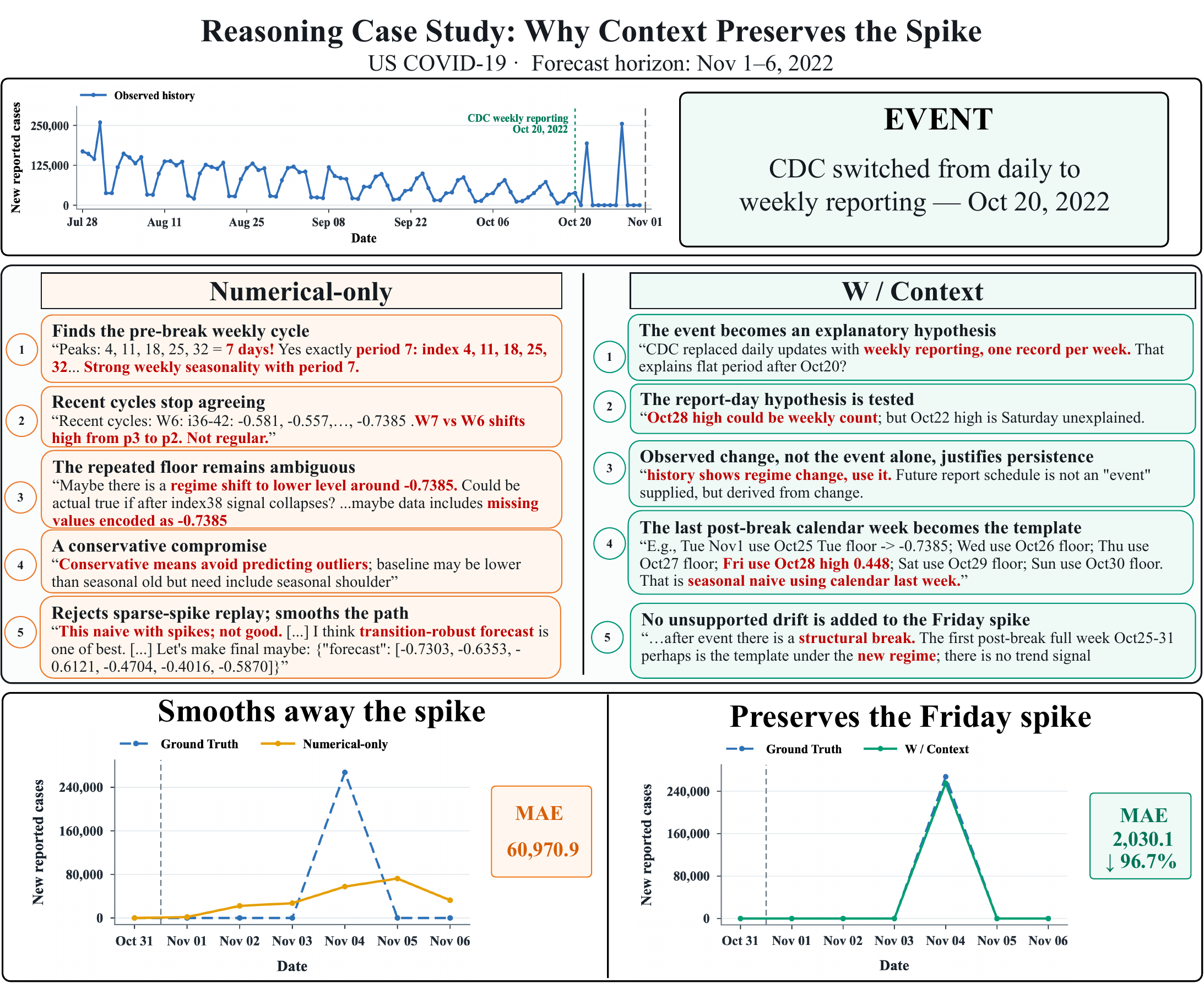}
    \end{minipage}\hfill
    \begin{minipage}[t]{0.49\linewidth}
        \centering
        \vspace{0pt}
        \includegraphics[width=\linewidth]{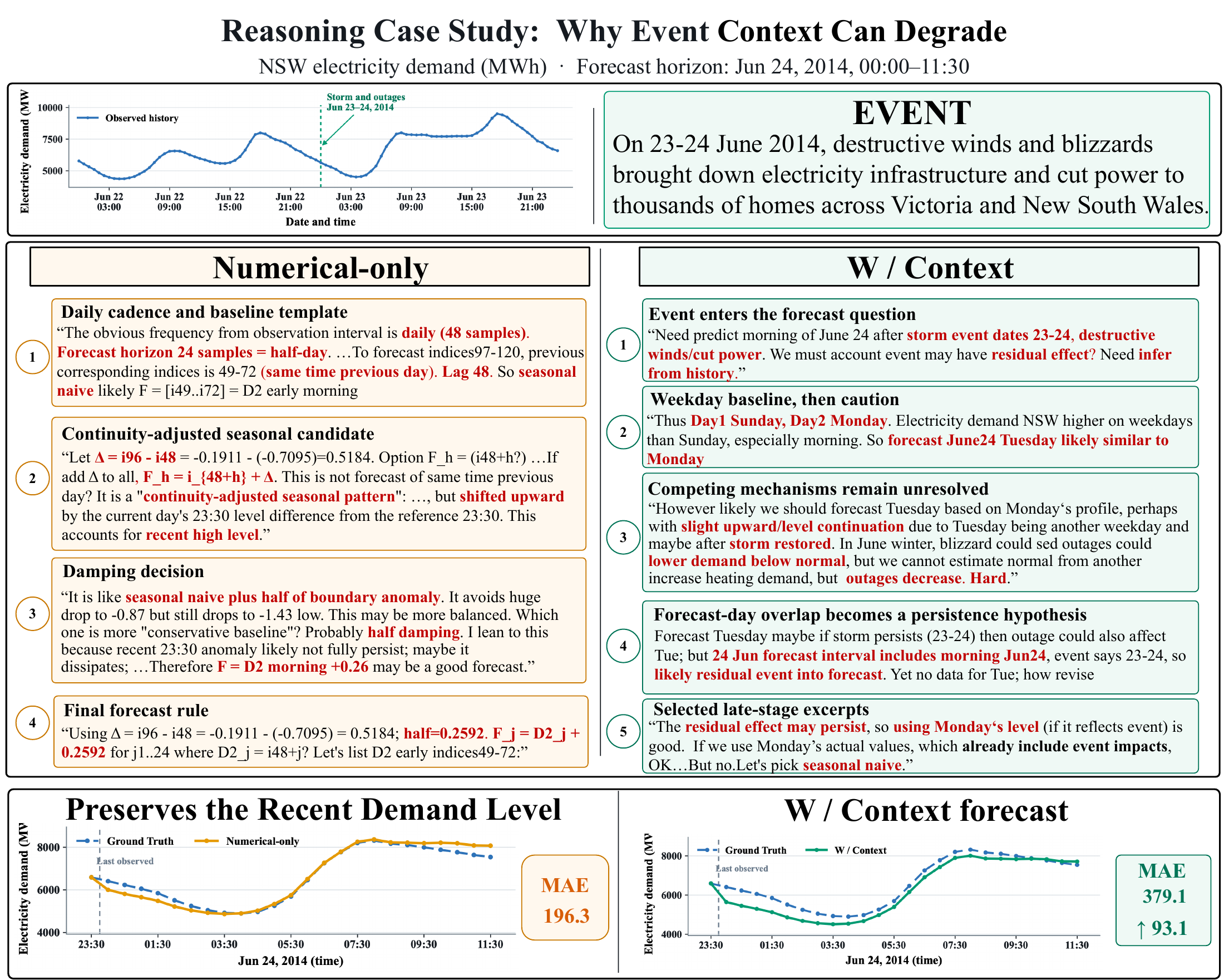}
    \end{minipage}
    \caption{Context-dependent reasoning in paired forecasting cases. Left: the reporting-change context helps preserve a Friday spike in the health forecast. Right: assuming that storm-related outages persist into the forecast window degrades the energy forecast.}
    \label{fig:reasoning_cases}
\end{figure}

To examine how context shapes individual forecasts beyond aggregate error, we compare the model's predictions and accompanying reasoning trajectories under identical numerical histories, where rationales reveal the model's stated use of evidence.
Figure~\ref{fig:reasoning_cases} presents one case in which context helps and another in which it hurts.
In the health case, the numerical-only model identifies an earlier weekly cycle yet treats the near-zero values after October~20 as ambiguous, smoothing away the subsequent Friday peak.
With context, it connects the pattern to the CDC's switch from daily to weekly reporting, checks the interpretation against the observed post-change week, and uses that week as a forecast template.
The Friday spike is retained and MAE falls by 96.7\%.

The energy case illustrates the opposite outcome.
The numerical-only forecast follows recent demand and its daily cycle, whereas the context-aware forecast assumes storm-related outages will continue to suppress demand.
Although the event plausibly explains earlier observations, extending its effect beyond the recent trajectory raises MAE by 93.1\%.
Together with Table~\ref{tab:forecasting_results}, these cases show that World2TS context helps in most evaluated comparisons, yet no individual contextual record is guaranteed to improve a forecast.
Determining when an event warrants revising a numerical prediction, and by how much, requires further exploration.
% \FloatBarrier

\section{Conclusion}
\label{sec:conclusion}

We introduced World2TS to connect heterogeneous time series with the real-world events that help explain their trajectories.
Its reusable construction tool, \texttt{world2ts-core}, combines agent-led semantic reasoning and adaptive evidence search with CLI-based execution and validation under human review.
The resulting \texttt{world2ts-data} spans 31 datasets across six domains, with 4,179 series, 97.1 million observations, 1,159 verified events, and 56,846 event--series links.
Our data analyses characterize the coverage and series-level relevance of these records, while controlled forecasting experiments show that adding World2TS context reduces error in 49 of 54 paired model--domain--metric comparisons.
Together, these findings establish World2TS as a reusable data foundation for developing and evaluating context-aware forecasting models and agents, and open the way to studying how real-world events can be integrated with numerical histories across domains.

\bibliographystyle{world2ts_scis}
\bibliography{world2ts_scis}

\end{document}